\documentclass[10pt,twocolumn,letterpaper]{article}

\usepackage{cvpr}
\usepackage{times}
\usepackage{epsfig}
\usepackage{graphicx}
\usepackage{amsmath}
\usepackage{amssymb}
\usepackage{authblk}

\def\Plus{\texttt{+}}

\usepackage[pagebackref=true,breaklinks=true,letterpaper=true,colorlinks,bookmarks=false]{hyperref}

\cvprfinalcopy % *** Uncomment this line for the final submission

\def\cvprPaperID{7} % *** Enter the CVPR Paper ID here

\ifcvprfinal\fi
\begin{document}
%%%%%%%%% TITLE
\title{Two Stream 3D Semantic Scene Completion}

\author[1]{Martin Garbade}
\author[2]{Yueh-Tung Chen}
\author[1]{Johann Sawatzky}
\author[1]{Juergen Gall}
% \author[2]{Author E\thanks{E.E@university.edu}}
\affil[1]{University of Bonn, \{garbade, sawatzky, gall\}@iai.uni-bonn.de }
\affil[2]{b-it, RWTH Aachen University, yuehtung.chen@gmail.com}

\maketitle
%\thispagestyle{empty}

%%%%%%%%% ABSTRACT
\begin{abstract}
Inferring the 3D geometry and the semantic meaning of surfaces, which are occluded, is a very challenging task. Recently, a first end-to-end learning approach has been proposed that completes a scene from a single depth image. The approach voxelizes the scene and predicts for each voxel if it is occupied and, if it is occupied, the semantic class label. In this work, we propose a two stream approach that leverages depth information and semantic information, which is inferred from the RGB image, for this task. The approach constructs an incomplete 3D semantic tensor, which uses a compact three-channel encoding for the inferred semantic information, and uses a 3D CNN to infer the complete 3D semantic tensor. In our experimental evaluation, we show that the proposed two stream approach substantially outperforms the state-of-the-art for semantic scene completion.                    

\end{abstract}

\section{Introduction}
Humans quickly infer the 3D semantics of a scene, i.e., an estimate of the 3D geometry and the semantic meaning of the surfaces. While RGB-D sensors in combination with CNNs provide geometry and semantic information, the resulting representation is very sparse since large parts of the 3D scene are occluded and not visible.  
The perception, however, is not limited to the visible part of the scene. When looking at a mug on a table, a human can estimate the full geometry of both objects including parts which are invisible since they are occluded by the objects themselves.
This information is obtained from semantic understanding of the scene which allows to estimate the spatial extent of the objects from experience. Such an ability is highly desirable for autonomous agents, e.g., to navigate or interact with objects. A robot that has an intuition about the geometry behind the surface it sees, for example, could plan ahead given a single view instead of exhaustively explore the occluded parts of a scene first.

In this work, we aim to estimate the semantics not only of the visible part, but of the entire scene including the occluded space. To this end, we build on the work of Song et al.~\cite{song2016ssc}. They show that semantic scene understanding and 3D scene completion benefit from each other. On one hand, recognizing a part of the object helps to estimate its location in the 3D space and the voxels it occupies. On the other hand, knowing the occupancy in the 3D space gives information on form and size of the object and thus facilitates semantic recognition. For estimating for each voxel in the scene the occupancy and semantic label, they proposed an end-to-end trainable 3D convolutional neural network (3D CNN) which incorporates context from a large field of view via dilated convolutions. The approach, however, only uses depth as input and neglects the RGB image. This means that the semantic label has to be inferred from the geometry alone and properties such as color, texture, or reflectance are not taken into account. 

We therefore extend the approach~\cite{song2016ssc} by keeping its beneficial context incorporation and end-to-end trainability while modifying it to leverage semantic information inferred from the RGB image at the input stage as well as at the loss. Given a single RGB-D image, we first use a 2D CNN to infer the semantic labels from the RGB data and construct an incomplete 3D semantic tensor. To this end, we map the inferred semantic labels to the 3D space and label each visible surface voxel by the inferred class label. The 3D semantic tensor is incomplete since it only contains the labels of the visible voxels but not of the occluded voxels. The 3D projection is performed using the depth image. The tensor is then used as input for a 3D CNN that infers a complete 3D semantic tensor, which includes the occupancy and semantic labels for all voxels.           

Using the RGB images as input leads to a significant performance gain in scene completion and semantic scene completion as our experiments show. We outperform \cite{song2016ssc} by a substantial margin of up to 9.4~$\%$ on NYU. This implies that RGB images provide a rich discriminative signal.

\section{Related Work}
Several works address the problem of semantic segmentation of RGB-D images, e.g., \cite{ren2012rgbd, lai2014unsupervised, gupta2013perceptual, Silberman:ECCV12}, but they infer semantic labels only for the visible pixels of the image, which means that occluded voxels are not reconstructed. 

A possible strategy for semantic scene completion is the generation of 3D object proposals and subsequent 3D shape completion of the respective object. The problem of completing the 3D shape has been addressed in several works~\cite{rock2015completing, nguyen2016field, varley2016shape, wu2015shapenets, wang2017shape, yang20173dobject, han2017high, dai2017shape}. In the context of inferring a voxel-wise segmentation, holes between objects have to be filled. As long as these missing parts are small, they can be filled using plane fitting \cite{monszpart2015rebuilding} or object symmetry \cite{kim2012acquiring, mattausch2014object}. Objects that are not detected, however, heavily disturb the 3D scene completion. Completing the scene geometry without predicting the semantics has been addressed by \cite{FirmanCVPR2016}. Their model assumes that objects of semantically dissimilar classes can still be represented by similar 3D shapes, i.e., it is possible to predict the unobserved voxels from the frontal geometry. However, this approach fails for complex scenes where these geometric constraints are violated.

An alternative is to use instances of 3D mesh models and fit them to the scene geometry \cite{Geiger2015GCPR, gupta2015aligning, kim2012acquiring, lai2010object, li2015object, nan2012search-classify, song2014sliding, shao2012interactive, li2015database, shi2016data-driven}. The mesh models, however, do not model shape variations of objects of the same category and increasing the number of mesh models per category is not practical since the number of available CAD models is limited and the shape retrieval becomes more expensive. The approaches \cite{jiang2013linear, lin2013holistic} even neglect fine-grained details and simply fit 3D primitives to the scene.

Various contextual cues proved to be helpful for semantic scene completion. While physical reasoning is employed in \cite{zheng2013beyond}, \cite{kim20133dscene} predict voxel labels with a conditional random field (CRF) whose unary potentials are determined by floor plans. The CRF, however, only models contextual information within a short distance. In \cite{blaha2016large} and \cite{hane2013joint}, semantic scene completion and multi-view reconstruction are jointly performed. The approaches do not rely on end-to-end learning approaches, but they use predefined features and heuristics to integrate context information. 

To facilitate learning of scene completion in an end-to-end manner, \cite{dai2017scannet, zhang2017physically, chang2017matterport3d}  collected large scale datasets with real world data. Previously, synthetic datasets were employed to provide ground truth data for object completion \cite{chang2015shapenet, wu2015shapenets} or for entire scenes \cite{handa2015scenenet}. Due to these datasets, training of an end-to-end approach for semantic scene completion became feasible~\cite{song2016ssc}. Parallel to our work, \cite{zhang18essc} have proposed a new network architecture which leverages sparse feature map encodings and allows for much deeper network architectures. While we address the problem of scene completion from a single viewpoint as in~\cite{song2016ssc, zhang18essc}, semantic scene completion from multiple RGB-D images is addressed in \cite{dai2018scancomplete}.

\section{Two Stream Semantic Scene Completion}
\subsection{Semantic Scene Completion}

\begin{figure*}[t]
\centering
  \includegraphics[width=12.0cm]{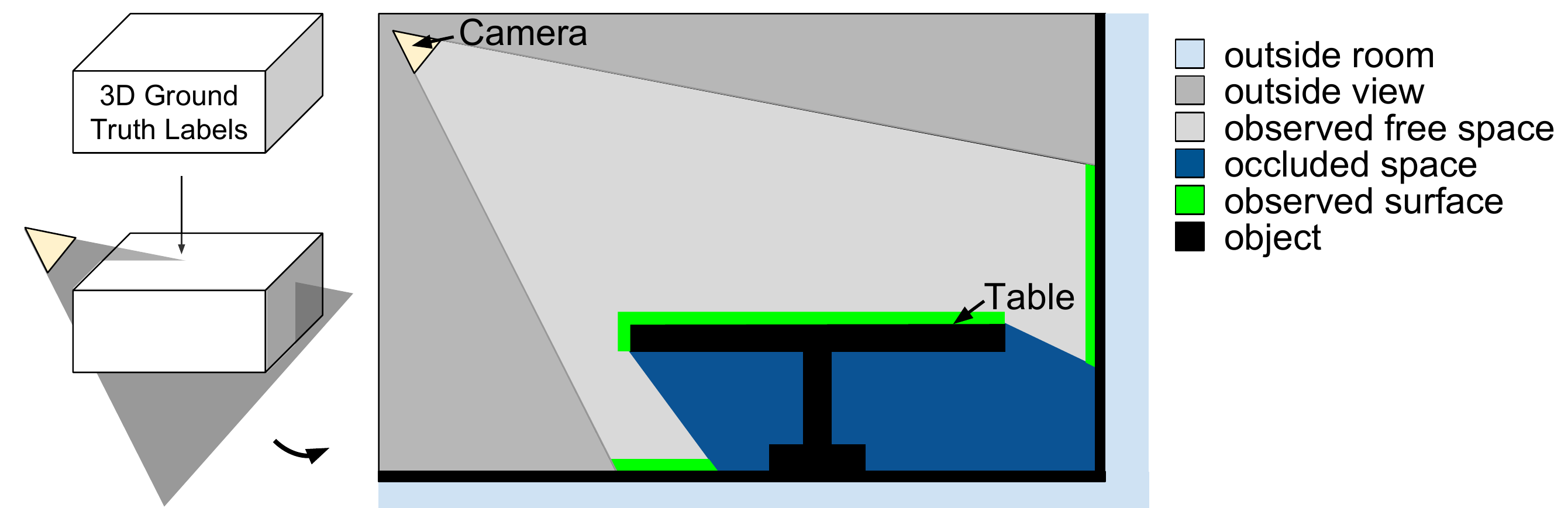}%
  \caption{Using the protocol of \cite{song2016ssc}, ground truth labels are provided for all voxels of a 3D volume. Voxels that are outside the intersection of the camera frustum and ground truth volume are outside the room or outside the view and not taken into account. Within the intersection, there are observed surface voxels (green) and observed non-occupied voxels (light gray), but other voxels are not observed by the camera. These voxels are either non-occupied (blue) or belong to an object (black).             
	}
\label{fig:scene_description}
\end{figure*}

The goal of 3D semantic scene completion is to classify every voxel in the view frustum into one of $K+1$ labels $c={c_0,...,c_{K}}$ where $c_0$ represents an empty voxel and $c_1,...,c_{K}$ represents one of $K = 11$ class labels like ceiling, floor, wall, window, chair, bed, sofa, table, tv, furniture and object. As illustrated in Figure~\ref{fig:scene_description}, the camera observes only a part of the scene while other voxels are occluded. The occluded voxels can either be empty ($c_0$) or belong to one of the $K$ classes.

\begin{figure*}[t]
\centering
  \includegraphics[width=12.0cm]{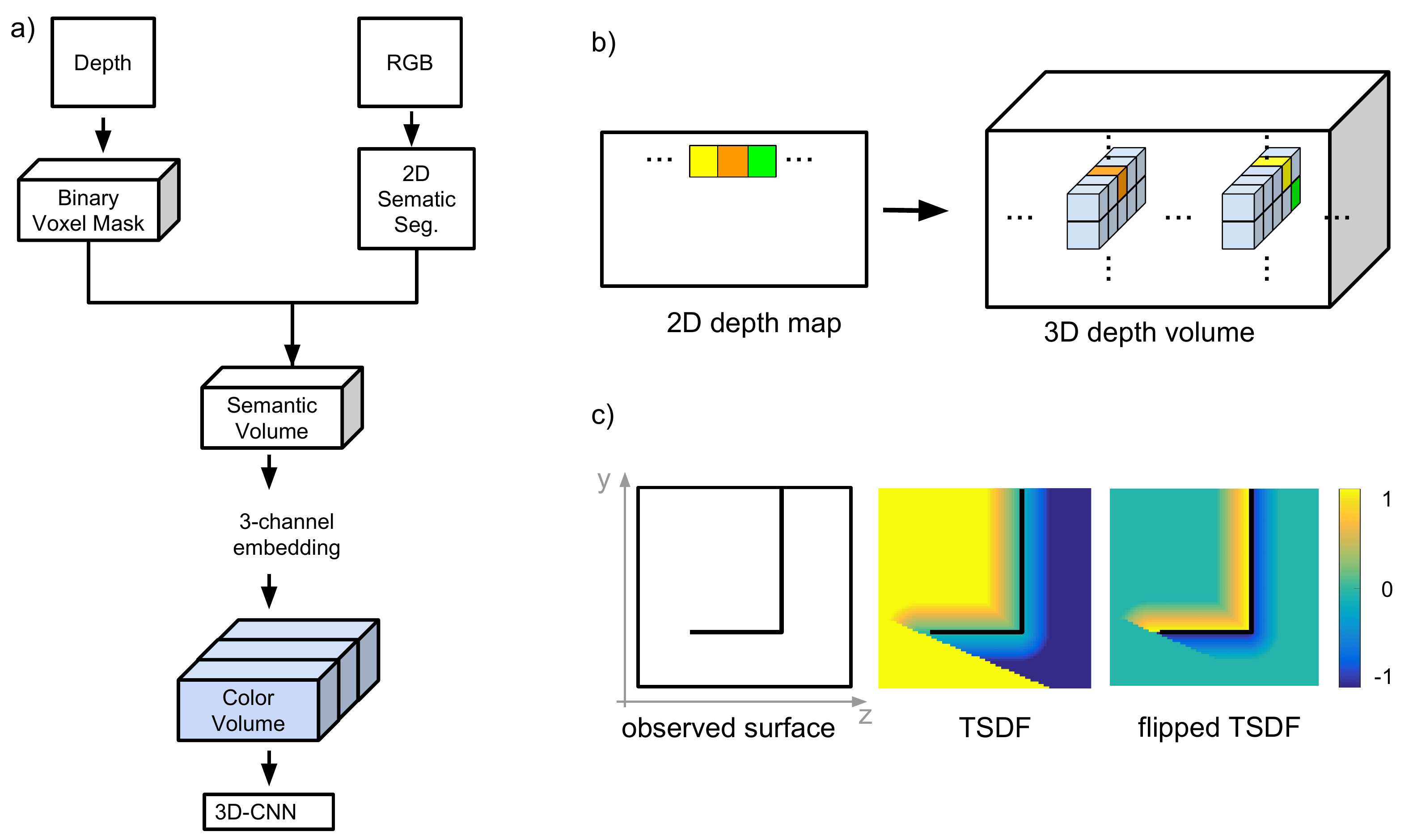}
  \caption{a) The proposed two stream approach for semantic scene completion transforms first the depth data and RGB image into a volumetric representation, which represents the geometry and semantic of the visible scene and then uses a 3D-CNN to infer a 3D semantic tensor for the entire scene. b) Given 2D depth map and camera pose, a binary voxel mask is created by setting each voxel that belongs to a depth pixel to one and all other voxels to zero (blue). c) Visualization of TSDF vs.\ flipped TSDF. One can see the long `shadow' caused by the observed surface which produces high gradients at the occlusion boundary (between -1 and 1). In the flipped TSDF, this effect is suppressed. The gradient is highest at the surface.
	}
\label{fig:project_depth}
\end{figure*}

To address the task of 3D semantic scene completion, we propose an approach that leverages two input streams, namely RGB and depth. An overview of the approach is given in Figure~\ref{fig:project_depth}~a). While the depth data is converted into a volumetric representation, the RGB image is first processed in a separate branch to infer 2D semantic segmentation maps and then transformed into a volumetric representation referred to as color-volume (Section~\ref{sec:color}). The volumetric representation is then fed to a 3D convolutional neural network (3D-CNN). The 3D-CNN infers a 3D semantic tensor where every voxel is classified as either being empty or belonging to one of the $K$ semantic classes. In the following, each step will be discussed in detail.

\subsection{Depth Input Stream}\label{sec:depth}

To obtain the volumetric input encoding, the depth map is projected into a regular voxel grid using the camera pose, which is provided along with each image. The voxel grid is of size 240 x 144 x 240 voxels and encodes a scene of 4.80m horizontally, 2.88m vertically, and 4.80m in depth with a resolution of 0.02m. For every pixel in the depth map, its corresponding voxel in the 3D input volume is computed using the camera pose. 
The obtained binary voxel mask encodes the location of surface points that are visible to the camera, see Figure \ref{fig:project_depth} b). 

As pre-processing, all 3D scenes are rotated such that the room orientations are aligned. For indoor room scenes, one can assume that most of the observed surface normals are oriented either like the normals of the walls, floor or ceiling, which are usually planar. Therefore a principal component analysis of the surface normals is used to infer the room orientation, which is used to align the scene.

\subsection{Color Input Stream}\label{sec:color}

The input RGB image is first processed by a 2D-CNN~\cite{DeepLabChen15}, which is an adaptation of the Resnet101 architecture \cite{ResNetHe16} for semantic segmentation. While all but one pooling layer are omitted, dilated convolutions are used to keep the output resolution high while simultaneously increasing the receptive field. The 2D-CNN predicts the softmax probabilities for every class and pixel at a resolution which is four times smaller than the input image. The output is then upsampled to the original resolution of the image using bilinear interpolation.  A densely connected CRF~\cite{CrfKoltun11} is then used in combination with the inferred class probabilities and the RGB image to refine the semantic segmentation map. For training, we use the same setting as in~\cite{DeepLabChen15}. As initialization, we use a model that is pre-trained on MSCOCO \cite{lin2014microsoft} and fine-tune it on the dataset for 3D semantic scene completion. 
Furthermore, we present results using the more recent model Deeplab v3\Plus \cite{deeplabv3plus2018} which is pretrained on ADE-20k and finetune it on NYUv2 using an initial learning rate of 0.001. We also apply a CRF~\cite{CrfKoltun11} on the resulting outputs.

As in Section \ref{sec:depth}, we convert the 2D segmentation map into a volumetric representation. Since each pixel in the depth map corresponds to a pixel in the 2D semantic segmentation map, every class pixel can be projected into the 3D volume at the location of its corresponding depth value. This yields an incomplete 3D semantic tensor that assigns to every surface voxel its corresponding class label. The class labels can be encoded by one-hot encoding, i.e., a channel for each class, or by a single channel for the class label. In our experiments, however, we show that none of them is optimal. Encoding semantic classes with only one channel implies a semantic proximity of classes by the numerical proximity of their class values, which introduces undesirable artifacts based on the class values. The one-hot encoding has the disadvantage that it is insufficient in terms of memory consumption since it requires to store a $K$ dimensional vector per voxel. We therefore represent the semantic information by a lower dimensional vector. We use a three-dimensional vector and encode the classes linearly from $(0,0,1)$ over $(0,1,1)$, $(0,1,0)$, $(1,1,0)$ to $(1,0,0)$.

\subsection{3D-CNN}\label{sec:3dcnn}

\begin{figure*}[t]
\centering
  \includegraphics[width=12.0cm]{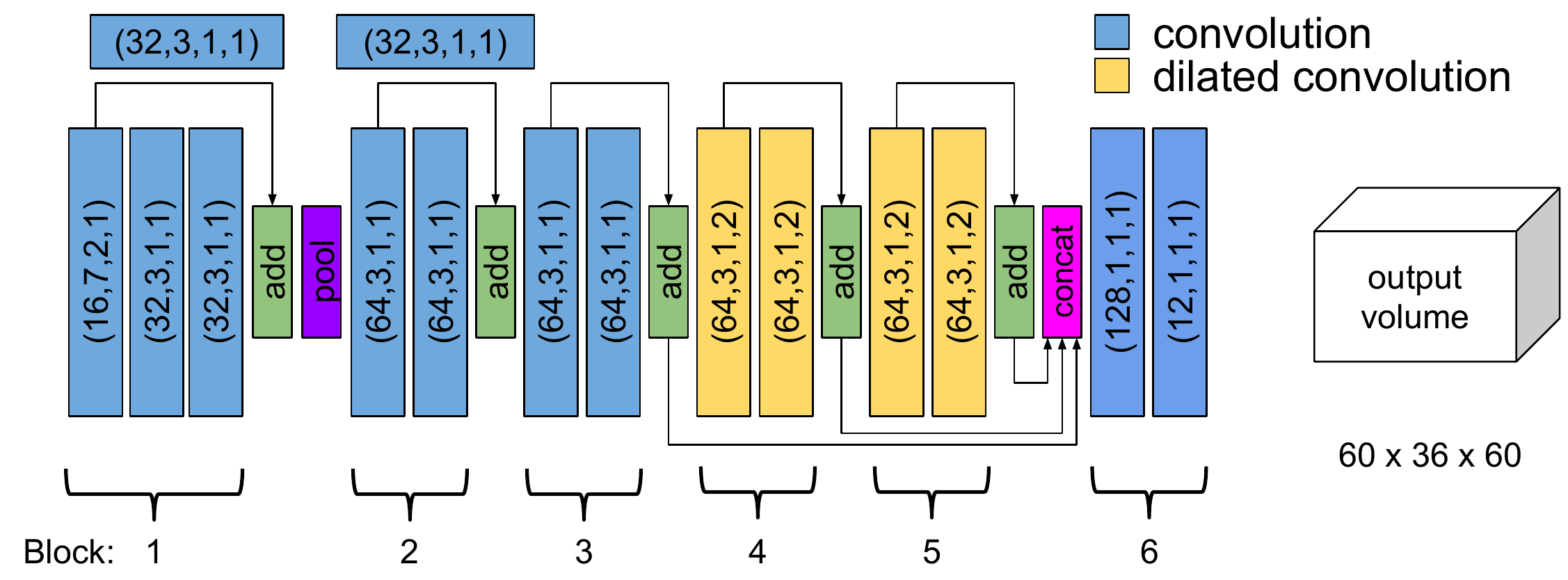}%
  \caption{Architecture of the 3D-CNN. The parameters of the convolution kernels are denoted as (number of filters, kernel size, stride, dilation). All but the last convolution layer have a ReLU activation function assigned to it. Arrows indicate skip connections \cite{ResNetHe16} where the output of one convolution layer is added to another output at a later stage. Pool denotes max pooling. The output is a volume that is 4 fold downsampled with respect to the input of the 3D CNN and encodes for every voxel the probability of it being empty (label 0) or to belong to one of $K$ semantic classes.
	}
\label{fig:sscnet}
\end{figure*}

For the 3D-CNN, we adapt the architecture of \cite{song2016ssc} by increasing the number of input channels of the first convolutional layer such that it fits to our input. The architecture is illustrated in Figure~\ref{fig:sscnet}. It is inspired by the 2D-CNN for semantic segmentation. The major difference apart from using 3D instead of 2D convolutions is that the network only has a depth of 14 convolutional layers. The network has therefore significantly less parameters than its two dimensional counter part. Moreover, batch-normalization layers are omitted due to the small size of the batches. 

We adapt the training protocol of \cite{song2016ssc} as follows.
We train for 150,000 steps with a learning rate of 0.01 that is reduced by a factor of 0.1 after 100,000 iterations. As optimizer, stochastic gradient (SGD) with momentum is applied. As initialization, we chose a random initialization with a Gaussian distribution with mean $\mu = 0$ and a standard deviation of $\sigma = 0.01$.

The output of the 3D-CNN is a semantic tensor of size 60 x 36 x 60 x $(K+1)$, where $K$ is the number of object classes and an additional class is added for empty voxels. We compute a softmax cross entropy loss on the unnormalized network outputs $y$:
    \begin{equation}\label{eq:soft}
  \mathcal{L} = - \sum_{i,c} {w}_{ic}  \hat{y}_{ic}  \log \left( \frac{e^{y_{ic}}}{\sum\limits_{c'\in\mathcal{C}} e^{y_{ic'}}} \right)
  \end{equation}
  where $\hat{y}_{ic}$ are the binary ground truth vectors, i.e., $\hat{y}_{ic}=1$ if voxel $i$ is labeled by class $c$, and ${w}_{ic}$ are the loss weights.
  Since the ratio of empty vs.\ occupied voxels is 9:1, the empty space is randomly subsampled. Therefore ${w}_{ic}$  is chosen as binary mask such that only 2N empty voxels are selected for loss calculation where N is the number of occupied voxels in the scene.

\section{Experimental Evaluation}
\label{sec:experiments}

\subsection{Evaluation Metric}
For evaluation, we follow the evaluation protocol of \cite{song2016ssc}, which evaluates the accuracy on a subset of voxels. 
%is used to evaluate the performance.
The evaluation considers only voxels that are part of the occluded space and within both the room and the field-of-view as shown in Figure \ref{fig:scene_description}. While generating the 3D semantic labels from the annotated CAD models, every voxel in the input volume is marked as being on surface, free space, occluded space, outside field of view, outside room or outside ceiling. For semantic scene completion, a binary evaluation mask is computed such that the evaluation metric is only computed for voxels which are either occluded, on surface or close to the surface (within the range of the TSDF function defined by \cite{song2016ssc}). 
For scene completion another mask is computed which comprises all voxels in the occluded space. To assess the quality of 3D scene completion, several metrics are computed. First we compute the Jaccard index, which measures the intersection over union (IoU) between ground truth and predicted voxel for every object category $c_1,...,c_{K}$. As an overall segmentation performance, we compute the average across all classes. For scene completion all voxels are considered to belong to one of the two classes empty vs.\ non-empty. All object categories $c_1,...,c_{K}$ are counted as `non-empty'. For completion, IoU as well as precision and recall are computed.

\subsection{Datasets}
We evaluate our method on the NYUv2 dataset, which is in the following denoted as NYU. NYU consists of indoor scenes that are captured via a Kinect sensor. For 3D semantic scene completion labels, we use the annotated 3D labels provided by \cite{rock2015completing}. They provide 1449 scenes, annotated with 11 classes, 795 of which are used for training and 654 for testing. These annotations consist of CAD models that are fitted into the scene. Since the CAD models do not exactly fit the shape of the annotated objects and neglect small objects such as clutter, there is a significant mismatch between the Kinect input data and the output labels. To address this problem, depth maps generated from the projections of the 3D annotations as in \cite{rock2015completing} are used for training. For evaluation, we consider two test sets. The first test set, which is denoted by NYU Kinect, consists of the depth maps from the Kinect sensor and the second test set, which is denoted as NYU CAD, uses the depth maps generated by projection.

\subsection{Ablation Study}
We conduct an ablation study on Kinect to analyze the design choices of our model.

\begin{table*}[ht]
\centering
\begin{tabular}{l | c|rrrrrrrrrrrrr}
					&	Scene Completion	& 	\multicolumn{12}{c}{Semantic Scene Completion}																							\\ \hline
semantic 					&	IoU	& 	ceil.	&	floor	&	wall	&	win.	&	chair	&	bed	&	sofa	&	table	&	tv	&	furn.	&	objs	&	avg	\\ \hline
DLv2						&	60.1	 & 	7.0	 & 	93.1	 & 	25.9	 & 	16.8	 & 	14.7	 & 	53.3	 & 	46.0	 & 	16.8	 & 	22.7	 & 	34.1	 & 	13.8	 & 	31.3	\\
DLv3\Plus					&	60.0	 & 	6.7	 & 	93.2	 & 	26.2	 & 	20.6	 & 	17.8	 & 	56.9	 & 	49.2	 & 	16.5	 & 	29.4	 & 	37.6	 & 	18.3	 & 	33.8	\\
GT						&	61.0	 & 	6.6	 & 	93.4	 & 	27.5	 & 	24.6	 & 	23.0	 & 	61.6	 & 	57.9	 & 	24.3	 & 	33.5	 & 	46.4	 & 	24.5	 & 	38.5 \\	
RGB image					&	58.2	 & 	4.2	 & 	93.4	 & 	19.3	 & 	4.4	 & 	10.8	 & 	34.9	 & 	20.2	 & 	11.8	 & 	4.9	 & 	17.2	 & 	10.3	 & 	21.0

\end{tabular}
\caption{Impact of the quality of the semantic input. For the version `RGB image', the 2D-CNN is omitted and the color values of the pixels instead of the semantic information is stored in the semantic volume. 
} 
\label{tab:semantic_input}
\end{table*}

\subsubsection{Effect of Semantic Input}
\begin{table*}[t]
\centering
\begin{tabular}{l | llllllllllll}
      & ceil. & floor & wall & win. & chair & bed  & sofa & table & tvs  & furn. & objs. & avg. \\ \hline
Deeplab v2  & 58.1  & 85.7  & 76.6 & 62.9 & 58.5  & 65.8 & 62.8 & 37.9  & 56.8 & 56.5  & 54.7  & 61.5 \\
Deeplab v3\Plus & 71.1  & 89.8  & 82.8 & 72.8 & 65.8  & 72.4 & 66.1 & 50.7  & 63.0 & 64.7  & 62.9  & 69.3
\end{tabular}
\caption{2D semantic segmentation accuracies on the NYUv2 dataset ($\% IoU$). In both cases, a CRF is used.} 
\label{tab:2D_semantic_segmentation}
\end{table*}

As mentioned in Section~\ref{sec:color}, we compare two network architectures for the 2D-CNN, namely Deeplab v2 (DLv2)~\cite{DeepLabChen15} and Deeplab v3\Plus\ (DLv3\Plus) \cite{deeplabv3plus2018}. Table~\ref{tab:semantic_input} shows that DLv3\Plus\ increases the accuracy for semantic scene completion from 31.3~$\%$ to 33.8~$\%$. This is expected since DLv3\Plus\ provides a better 2D segmentation accuracy compared to DLv2 as shown in Table~\ref{tab:2D_semantic_segmentation}. For scene completion, IoU is slightly higher for DLv2 than for DLv3\Plus.

We compare the results to a setting when we use ground truth semantic segmentation masks for the RGB images as input to the 3D-CNN, which is denoted by GT in Table~\ref{tab:semantic_input}. This also serves as an upper bound for our method when the used 2D-CNN provides perfect segmentation masks. As expected, using ground truth segmentation masks improves the semantic scene completion compared to DLv3\Plus\ by \Plus4.7~$\%$. For scene completion, the improvements compared to DLv3\Plus\ are \Plus1.0~$\%$. This shows that the quality of the 2D-CNN has a strong impact on the accuracy of semantic scene completion but only a minor impact on scene completion, which is also not the focus of this work.

As it is illustrated in Figure~\ref{fig:project_depth}~a), the RGB images are processed by a 2D-CNN and the inferred pixel-wise labels are used to construct the semantic volume. We also evaluate in Table~\ref{tab:semantic_input} what happens if the three-channel encoding is not based on the semantic labels but if the RGB values of the pixels are directly used for the encoding, i.e., without inferring semantic information from the visible part of the scene. This setting is denoted by `RGB image' and performs as expected poorly. Besides of the class `floor', all categories are poorly estimated.

\begin{table*}[ht]
\centering

\begin{tabular}{l | c|rrrrrrrrrrrrr}
					&	Scene Completion	& 	\multicolumn{12}{c}{Semantic Scene Completion}																							\\ \hline
channels					&	IoU	& 	ceil.	&	floor	&	wall	&	win.	&	chair	&	bed	&	sofa	&	table	&	tv	&	furn.	&	objs	&	avg	\\ \hline
1						&	59.3	 & 	8.3	 & 	93.3	 & 	25.0	 & 	13.4	 & 	11.5	 & 	43.0	 & 	31.8	 & 	11.2	 & 	2.4	 & 	26.5	 & 	16.8	 & 	25.8	\\
3						&	60.0	 & 	6.7	 & 	93.2	 & 	26.2	 & 	20.6	 & 	17.8	 & 	56.9	 & 	49.2	 & 	16.5	 & 	29.4	 & 	37.6	 & 	18.3	 & 	33.8	\\
12 (one-hot)					&	60.0	 & 	9.7	 & 	93.4	 & 	25.5	 & 	21.0	 & 	17.4	 & 	55.9	 & 	49.2	 & 	17.0	 & 	27.5	 & 	39.4	 & 	19.3	 & 	34.1	

\end{tabular}
\caption{Impact of the number of channels for the semantic volume. 12 channels refers to one-hot encoding. 
} 
\label{tab:input_encoding}
\end{table*}

\begin{table*}[ht]
\centering

\resizebox{0.80\textwidth}{!}{
\begin{minipage}{\textwidth}
% \begin{tabular}{l | c|ccccccccccccc}
\begin{tabular}{l | c|rrrrrrrrrrrrr}
					&	Scene Completion	& 	\multicolumn{12}{c}{Semantic Scene Completion}																							\\ \hline
input					&	IoU	& 	ceil.	&	floor	&	wall	&	win.	&	chair	&	bed	&	sofa	&	table	&	tv	&	furn.	&	objs	&	avg	\\ \hline
proposed, no fTSDF					&	60.0	 & 	6.7	 & 	93.2	 & 	26.2	 & 	20.6	 & 	17.8	 & 	56.9	 & 	49.2	 & 	16.5	 & 	29.4	 & 	37.6	 & 	18.3	 & 	33.8	\\
with fTSDF, early fusion				&	60.2	 & 	8.1	 & 	94.4	 & 	25.6	 & 	17.1	 & 	17.8	 & 	53.6	 & 	48.0	 & 	17.0	 & 	28.0	 & 	36.0	 & 	18.4	 & 	33.1	\\
with fTSDF, fusion 1					&	60.0	 & 	4.8	 & 	94.1	 & 	25.5	 & 	21.5	 & 	16.6	 & 	56.9	 & 	47.2	 & 	16.7	 & 	27.5	 & 	37.3	 & 	18.1	 & 	33.3	\\
with fTSDF, fusion 2					&	54.4	 & 	5.9	 & 	93.6	 & 	22.0	 & 	11.0	 & 	16.5	 & 	50.4	 & 	41.0	 & 	12.4	 & 	23.1	 & 	31.9	 & 	12.4	 & 	29.1	\\
with fTSDF, fusion 5					&	59.1	 & 	5.1	 & 	92.9	 & 	23.0	 & 	19.4	 & 	15.1	 & 	53.9	 & 	46.7	 & 	16.3	 & 	28.2	 & 	34.6	 & 	15.0	 & 	31.8	\\
with fTSDF, late fusion					&	60.4	 & 	5.7	 & 	93.9	 & 	25.7	 & 	20.3	 & 	15.9	 & 	55.7	 & 	44.8	 & 	17.0	 & 	28.1	 & 	34.9	 & 	16.0	 & 	32.5	\\
RGB image						&	58.2	 & 	4.2	 & 	93.4	 & 	19.3	 & 	4.4	 & 	10.8	 & 	34.9	 & 	20.2	 & 	11.8	 & 	4.9	 & 	17.2	 & 	10.3	 & 	21.0

\end{tabular}

\caption{Impact of the input for the 3D-CNN. The proposed architecture is shown in Figure~\ref{fig:project_depth}~a). The versions `with fTSDF' refers to a version where not only the semantic volume but also the flipped TSDF volume \cite{song2016ssc} are used.       
} 
\label{tab:fusion_schemes}
\end{minipage}}
\end{table*}

\subsubsection{Input Encoding}
As we discussed in Section~\ref{sec:color}, the encoding of the semantic information in the semantic volume should provide a numerical equidistance between classes, which can be achieved by using one-hot encoding. However, this approach has a high memory footprint. As an alternative, we evaluate a one-channel and a three-channel input encoding. In the one-channel setup, the numeric class values are normalized to the range from 0 to 1. For the proposed 3-channel input encoding, every label is mapped to a 3 dimensional vector as described in Section~\ref{sec:color}. Table~\ref{tab:input_encoding} shows that using only one channel performs poorly since it introduces undesirable artifacts based on the class values. While some classes like `floor' and `bed' are well recognized, the accuracy for `window' and `tv' is very low. Using one-hot encoding (12 channels) performs much better than 1 channel but it is expensive in terms of memory consumption. The proposed three-channel encoding requires less memory while it only slightly decreases the accuracy. Also the training time of the 3 channel setup is by a factor of 1.7 faster which reduces the training time from 4 to 2.5 days. Therefore we adopt the 3 channel setup as it provides an efficient alternative to the one-hot encoding.

\subsubsection{Fusion with flipped TSDF}
Furthermore, we have conducted an experiment where we combine our input with the flipped truncated signed distance function (fTSDF) proposed by \cite{song2016ssc} and evaluate different fusing schemes.

The fTSDF is computed as follows: The previously computed binary voxel mask (Figure \ref{fig:project_depth} b) is used to first compute a truncated signed distance function (TSDF) encoding as illustrated in Figure \ref{fig:project_depth} c). In the TSDF, every voxel contains as value the distance $d$ to the next surface point. The sign of the distance value indicates whether a voxel lies in the empty (1) or occluded space (-1). The TSDF has the disadvantage of having high gradients at the occlusion boundary, i.e., the boundary between observed and unobserved space behind a surface. Therefore in the TSDF encoding every surface yields a shadow into the unobserved space as shown in Figure \ref{fig:project_depth} c).

To provide a more meaningful input signal, the signed distance function is transformed into a flipped TSDF \cite{song2016ssc}, where every signed distance value $d$ is converted into a distance $d_f$ which is 1 or -1 at a surface and linearly falls to 0 at a distance $d_{max}$ from the surface: 
  \begin{equation}\label{eq:flippedTsdf}
d_f = sign(d) \mathcal{H}(d_{max} - \vert d \vert)  \frac{d_{max} - \vert d \vert}{d_{max}}
  \end{equation}
  where $d_{max}$ is the maximum distance of 24 cm and $\mathcal{H}$ is the Heaviside function:
\begin{equation}
\mathcal{H}(x) =
  \begin{cases}
  1 & \text{if } x\geq0\\
  0 & \text{if } x<0.
  \end{cases}
\end{equation}

We perform different fusion experiments to evaluate whether the proposed fTSDF encoding can give us a meaningful signal for semantic scene completion. As one can see from Figure~\ref{fig:sscnet} the 3D-CNN consists of several blocks. We concatenate the fTSDF before block 1 (early fusion) and also after block 1, 2 and 5 (fusion 1, 2 and 5). Before concatenation, both the color and the fTSDF input stream are processed separately. In the case of ``late fusion'' we take the maximum of the softmax probabilities of both streams.

As can be seen from Table~\ref{tab:fusion_schemes}, all fusion schemes perform slightly worse than our approach. This indicates that fTSDF provides a superfluous signal for our approach. This is interesting since computing the flipped TSDF volume is the most time-consuming part for inference and our approach provides a substantial faster alternative while also increasing the accuracy.          
%The computation of the flipped TSDF takes $\sim$7 seconds on a Nvidia Titan X Pascal.

\subsection{Comparison to State-of-the-Art}
\begin{table*}[t]

\resizebox{0.80\textwidth}{!}{
\begin{minipage}{\textwidth}
% \begin{tabular}{ l  l  | c | c c c c c c c c c c c c }
\begin{tabular}{ l  l  | c | r r r r r r r r r r r r}

NYU CAD	&				&	Scene Completion	& 	\multicolumn{12}{c}{Semantic Scene Completion}																							\\ \hline
method	&	trained on			&	IoU	&	ceil.	&	floor	&	wall	&	win.	&	chair	&	bed	&	sofa	&	table	&	tv	&	furn.	&	objs	&	avg	\\ \hline
Zheng et al. \cite{zheng2013beyond}	&	NYU			&	34.6	&		&		&		&		&		&		&		&		&		&		&		&		\\
Firman et al. \cite{FirmanCVPR2016}	&	NYU			&	50.8	&		&		&		&		&		&		&		&		&		&		&		&		\\
SSCNet \cite{song2016ssc}	&	NYU			&	70.3	&		&		&		&		&		&		&		&		&		&		&		&		\\
SSCNet \cite{song2016ssc}	&	SUNCG+NYU			&	73.2	&	32.5	&	92.6	&	49.2	&	8.9	&	33.9	&	57.0	&	59.5	&	28.3	&	8.1	&	44.8	&	25.1	&	40.0	\\
Ours, 3ch	&	NYU			&	76.1	 & 	28.3	 & 	94.0	 & 	48.6	 & 	33.0	 & 	33.4	 & 	67.9	 & 	54.7	 & 	31.1	 & 	33.8	 & 	50.8	 & 	30.6	 & 	46.0	\\
Ours, one-hot	&	NYU			&	76.1	 & 	25.9	 & 	93.8	 & 	48.9	 & 	33.4	 & 	31.2	 & 	66.1	 & 	56.4	 & 	31.6	 & 	38.5	 & 	51.4	 & 	30.8	 & 	46.2	\\
\multicolumn{15}{c}{}																															\\
NYU Kinect	&				&	Scene Completion	& 	\multicolumn{12}{c}{Semantic Scene Completion}																							\\ \hline
Lin et al. \cite{lin2013holistic}	&	NYU			&	36.4	&	0.0	&	11.7	&	13.3	&	14.1	&	9.4	&	29.0	&	24.0	&	6.0	&	7.0	&	16.2	&	1.1	&	12.0	\\
Geiger et al.\cite{Geiger2015GCPR}	&	NYU			&	44.4	&	10.2	&	62.5	&	19.1	&	5.8	&	8.5	&	40.6	&	27.7	&	7.0	&	6.0	&	22.6	&	5.9	&	19.6	\\
SSCNet\cite{song2016ssc}	&	NYU			&	55.1	&	15.1	&	94.7	&	24.4	&	0.0	&	12.6	&	32.1	&	35.0	&	13.0	&	7.8	&	27.1	&	10.1	&	24.7	\\
SSCNet\cite{song2016ssc}	&	SUNCG+NYU			&	56.6	&	15.1	&	94.6	&	24.7	&	10.8	&	17.3	&	53.2	&	45.9	&	15.9	&	13.9	&	31.1	&	12.6	&	30.5	\\
ESSCN \cite{zhang18essc}	&	NYU			&	56.2	&	17.5	&	75.4	&	25.8	&	6.7	&	15.3	&	53.8	&	42.4	&	11.2	&	0.0	&	33.4	&	11.8	&	26.7	\\
Ours, 3ch	&	NYU			&	60.0	 & 	6.7	 & 	93.2	 & 	26.2	 & 	20.6	 & 	17.8	 & 	56.9	 & 	49.2	 & 	16.5	 & 	29.4	 & 	37.6	 & 	18.3	 & 	33.8	\\
Ours, one-hot	&	NYU			&	60.0	 & 	9.7	 & 	93.4	 & 	25.5	 & 	21.0	 & 	17.4	 & 	55.9	 & 	49.2	 & 	17.0	 & 	27.5	 & 	39.4	 & 	19.3	 & 	34.1	

\end{tabular}
\caption{Comparison to the state-of-the-art.} 
\label{tab:results_state_of_art}
\end{minipage}}
\end{table*}

We evaluate our approach on the two test sets NYU CAD and NYU Kinect, which are in the following denoted as CAD and Kinect, and we compare our approach to the state-of-the-art. The results for scene completion and semantic scene completion are reported in Table~\ref{tab:results_state_of_art}. Both of our approaches with 3-channel input encoding and one-hot encoding perform comparably. Since one-hot encoding yields a slightly higher accuracy, we only discuss the difference between the latter and other approaches from the literature.

Our approach sets the new state-of-the-art for semantic scene completion. We achieve 46.2~$\%$ on CAD and 34.1~$\%$ on Kinect and outperform the approach by Song et al.~\cite{song2016ssc} by \Plus6.2~$\%$ on CAD and \Plus3.6~$\%$ on Kinect, although they use SUNCG as additional training data. If the same training data, i.e. only NYU, is used, our approach outperforms \cite{song2016ssc} by \Plus9.4~$\%$ on Kinect. For scene completion, we outperform \cite{song2016ssc} by \Plus5.8~$\%$ on CAD and \Plus4.9~$\%$ on Kinect if NYU is used as training data. However, even if \cite{song2016ssc} uses additional training data from SUNCG, our approach still outperforms it by \Plus2.9~$\%$ on CAD and \Plus3.4~$\%$ on Kinect. 

Compared to the recent ESSCN approach \cite{zhang18essc}, we perform better in both scene completion (\Plus3.8~$\%$) and semantic scene completion (\Plus7.4~$\%$). 
Note also that pretraining on SUNCG \cite{song2016ssc} and using a stronger 3D-CNN architecture \cite{zhang18essc} are orthogonal to our proposed method. One can assume that our performance would further increase by incorporating both ideas.

Table~\ref{tab:results_state_of_art} also includes the results of other approaches that do not rely on end-to-end learning~\cite{zheng2013beyond,FirmanCVPR2016,lin2013holistic,Geiger2015GCPR}. Furthermore, the approaches \cite{zheng2013beyond} and \cite{FirmanCVPR2016} only address scene completion but not semantic scene completion. These methods perform substantially worse than the end-to-end learning approaches.

\begin{figure*}[ht]
  \centering

  \includegraphics[width=1.8cm]{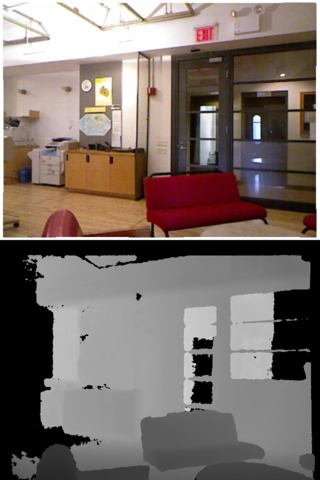}%
  \hspace{0.3cm}%    
  \includegraphics[width=3.8cm]{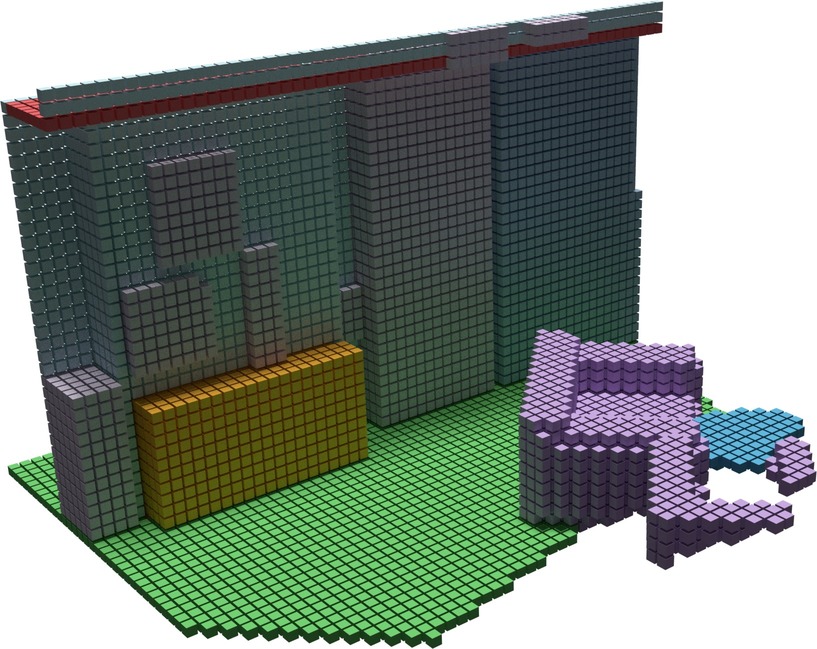}%
  \hspace{0.3cm}%
  \includegraphics[width=3.8cm]{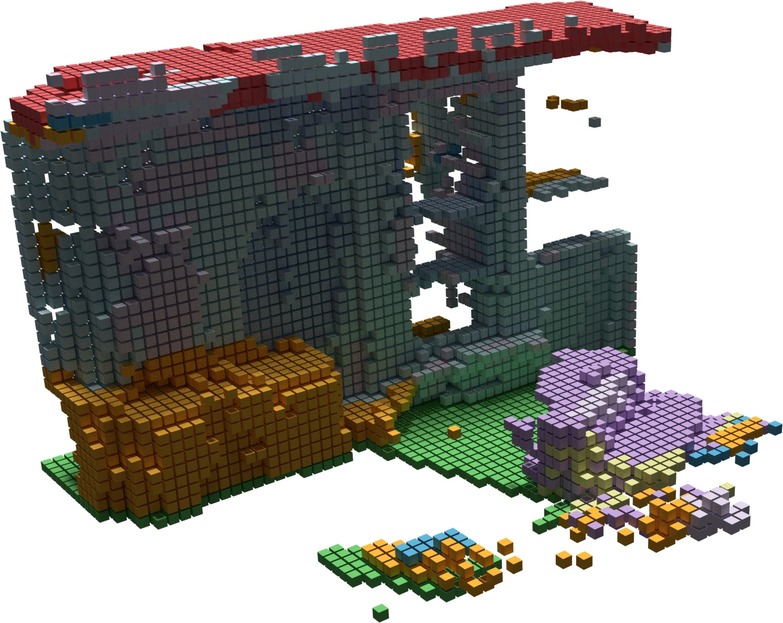}%
  \hspace{0.3cm}%  
    \includegraphics[width=3.8cm]{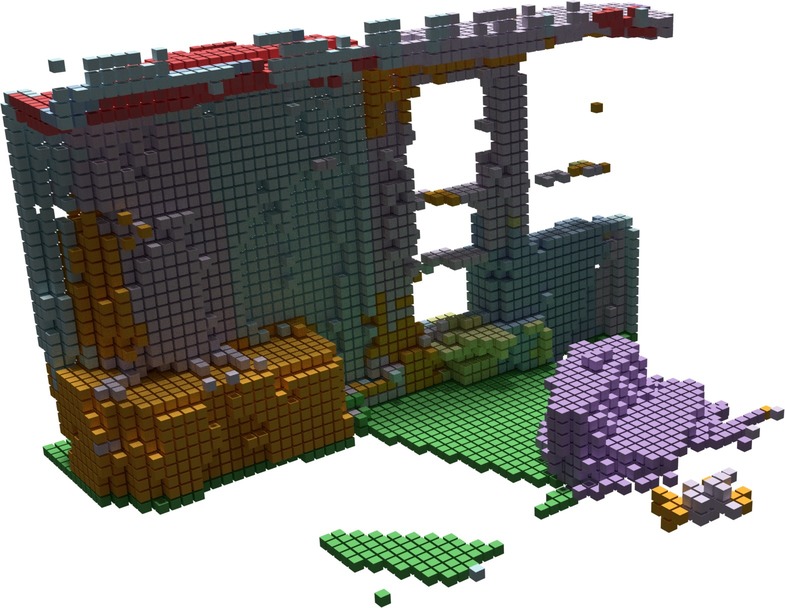}\\ \vspace{.25cm}%

  \includegraphics[width=1.8cm]{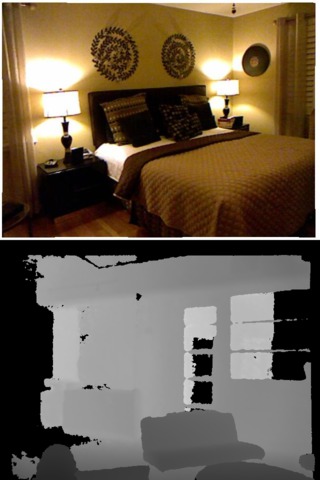}%
  \hspace{0.3cm}%
  \includegraphics[width=3.8cm]{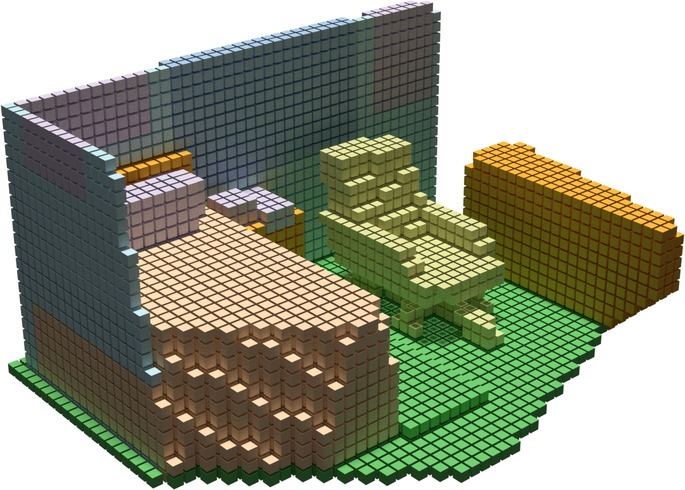}%
  \hspace{0.3cm}%
  \includegraphics[width=3.8cm]{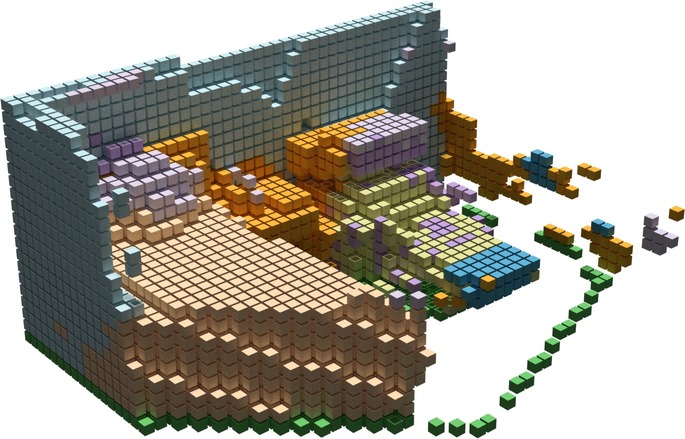}%
  \hspace{0.3cm}%  
    \includegraphics[width=3.8cm]{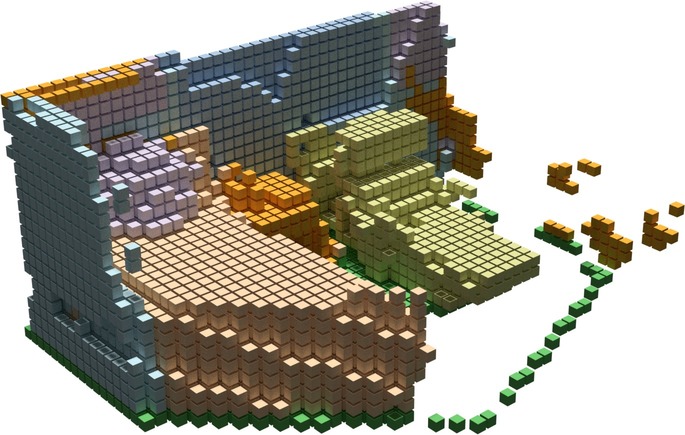}\\ \vspace{.25cm}%

  \includegraphics[width=1.8cm]{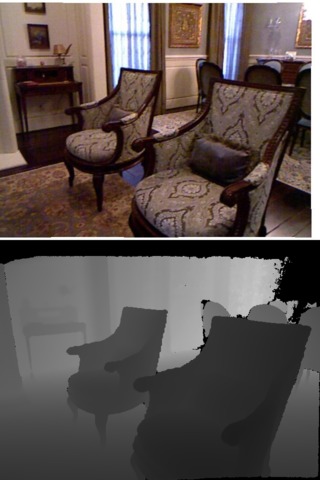}%
  \hspace{0.3cm}%  
  \includegraphics[width=3.8cm]{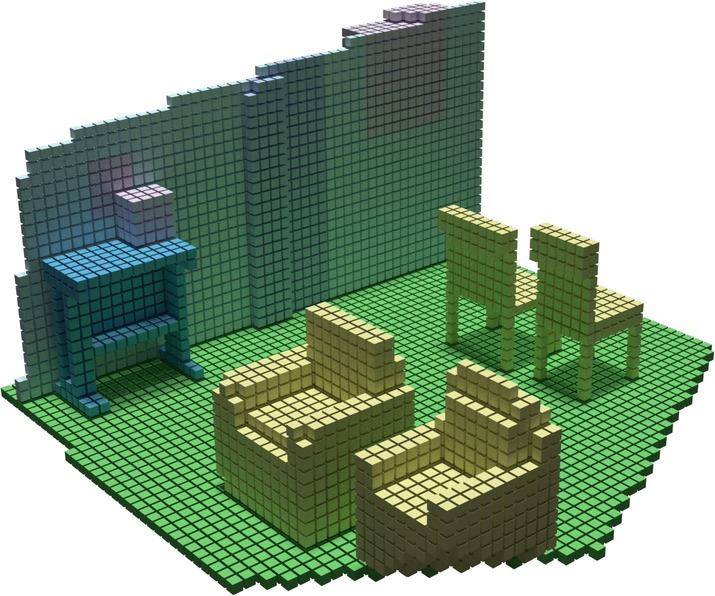}%
  \hspace{0.3cm}%
  \includegraphics[width=3.8cm]{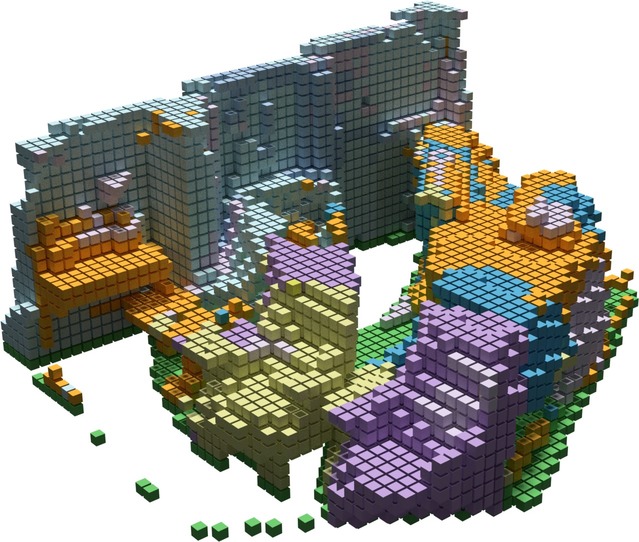}%
  \hspace{0.3cm}%  
    \includegraphics[width=3.8cm]{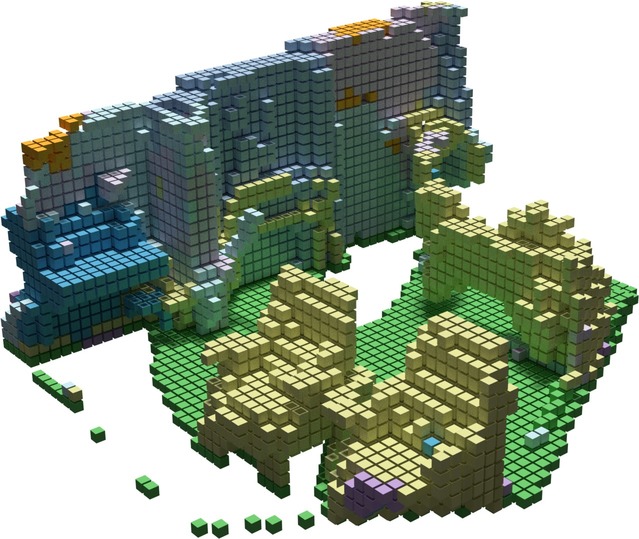}\\ \vspace{.25cm}%

  \includegraphics[width=1.8cm]{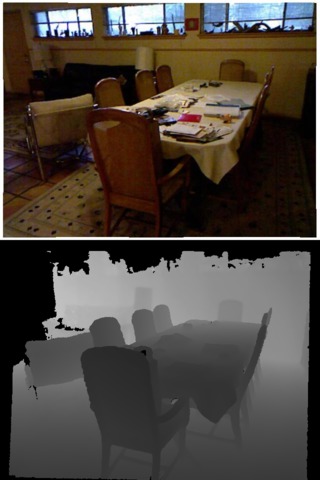}%
  \hspace{0.3cm}%    
  \includegraphics[width=3.8cm]{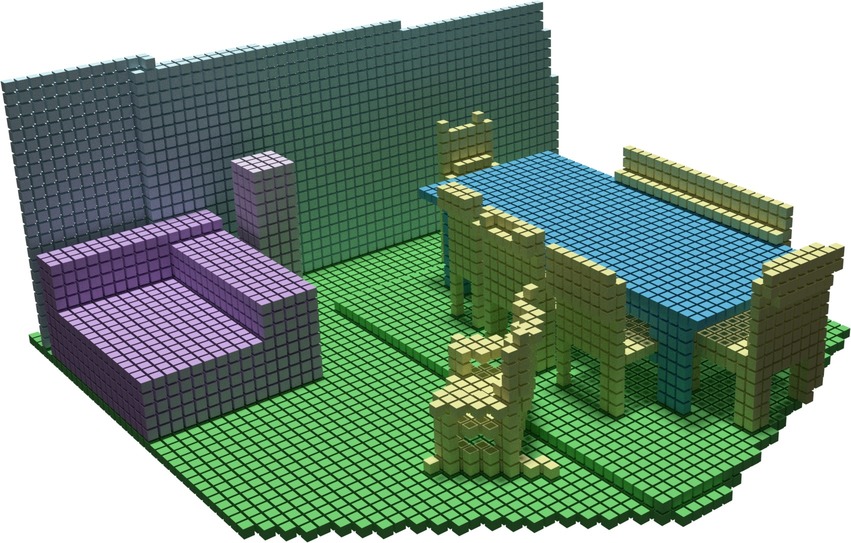}%
  \hspace{0.3cm}%
  \includegraphics[width=3.8cm]{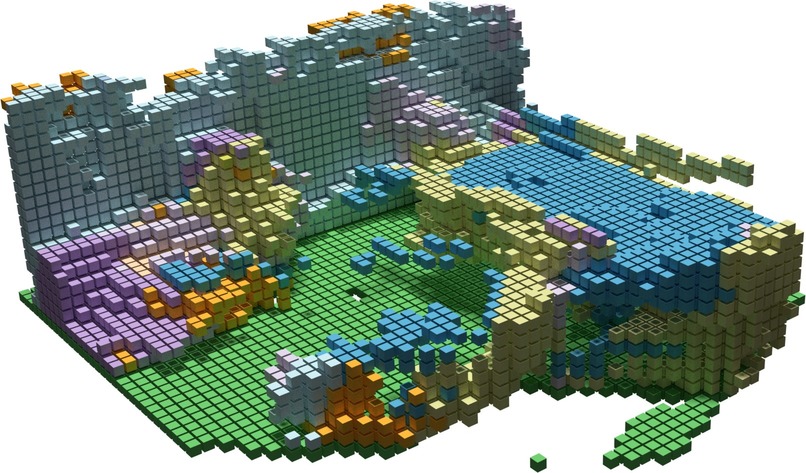}%
  \hspace{0.3cm}%  
    \includegraphics[width=3.8cm]{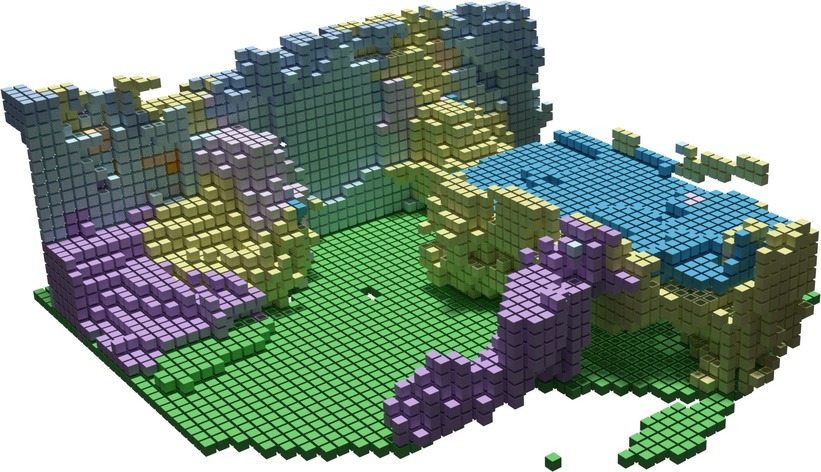}\\ \vspace{.25cm}%

  \includegraphics[width=1.8cm]{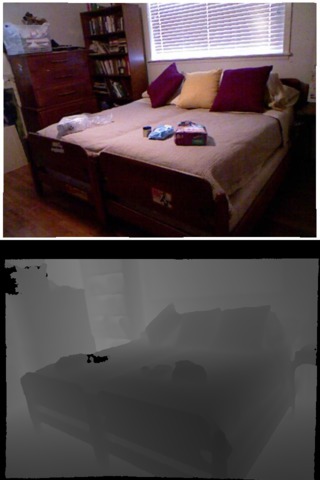}%
  \hspace{0.3cm}%    
  \includegraphics[width=3.8cm]{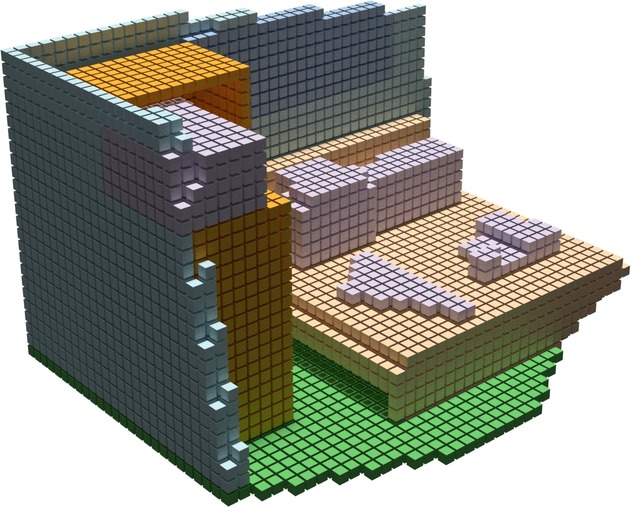}%
  \hspace{0.3cm}%
  \includegraphics[width=3.8cm]{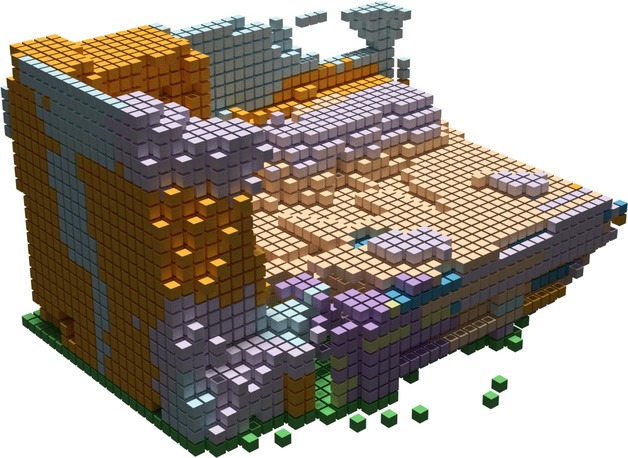}%
  \hspace{0.3cm}%  
    \includegraphics[width=3.8cm]{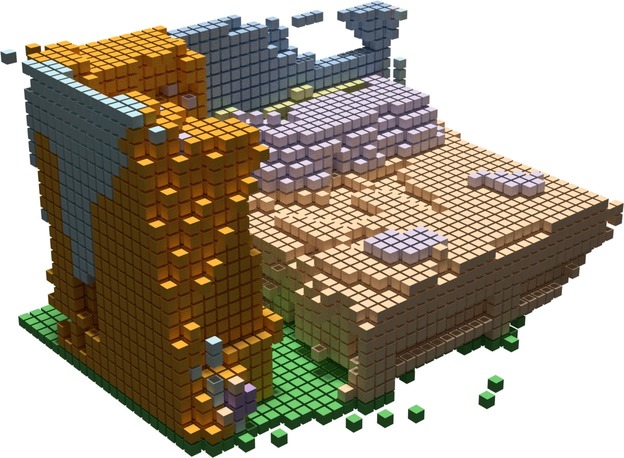}\\ \vspace{.25cm}%    

  \includegraphics[width=1.8cm]{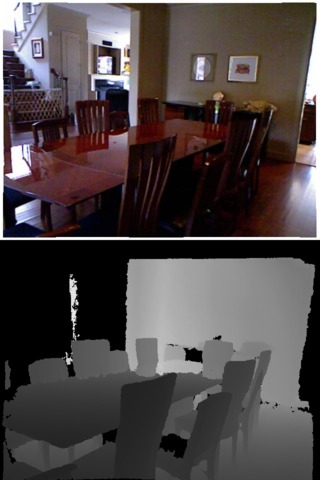}%
  \hspace{0.3cm}%    
  \includegraphics[width=3.8cm]{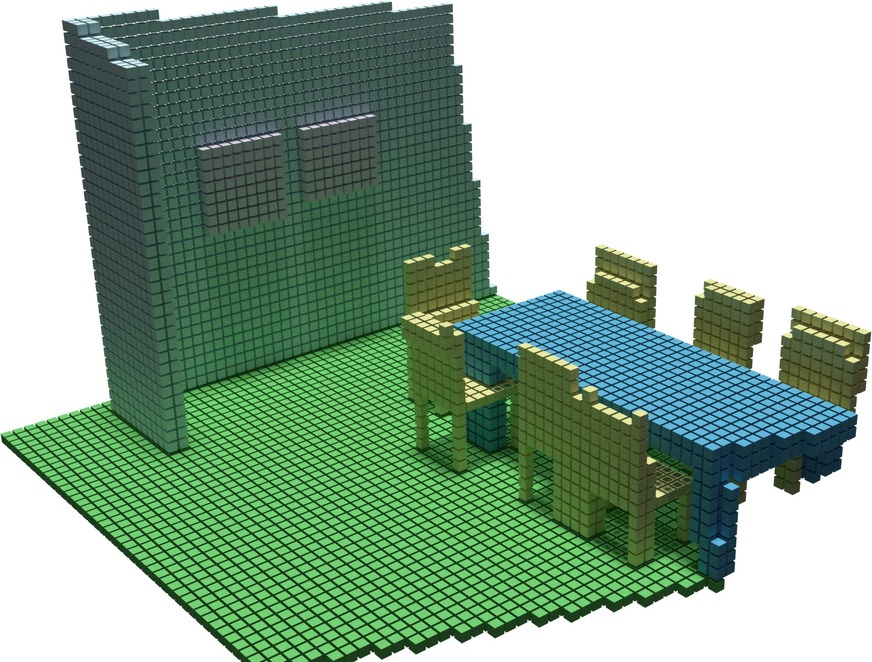}%
  \hspace{0.3cm}%
  \includegraphics[width=3.8cm]{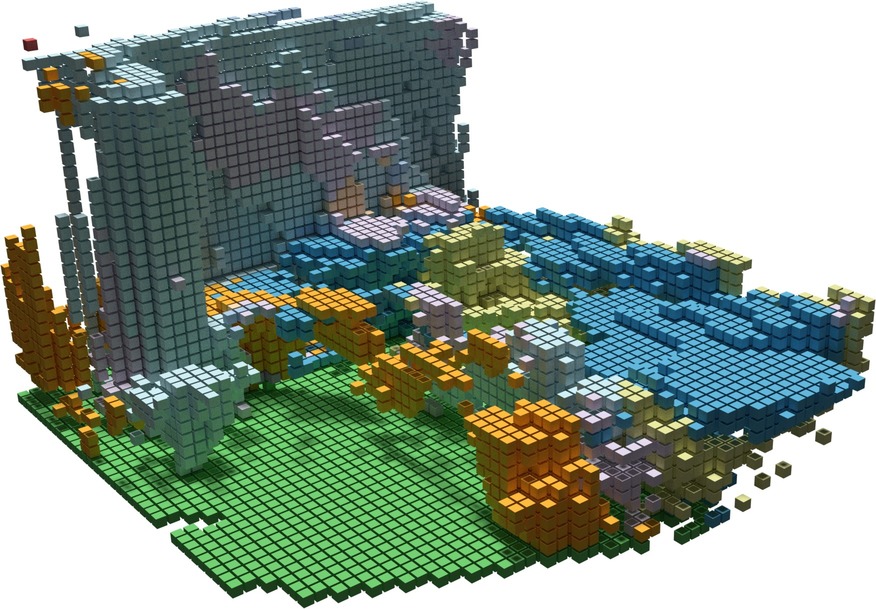}%
  \hspace{0.3cm}%  
    \includegraphics[width=3.8cm]{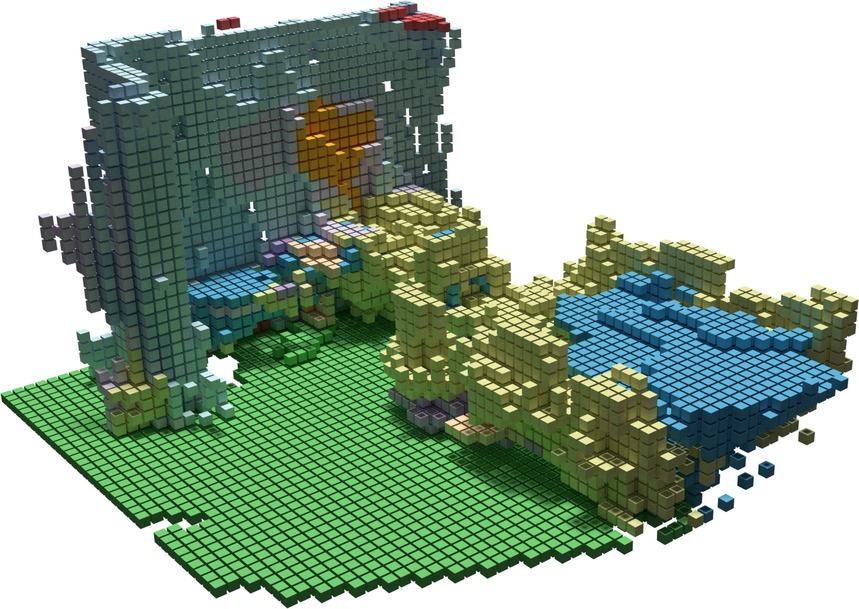}\\ \vspace{.25cm}%        

  \includegraphics[width=8.8cm]{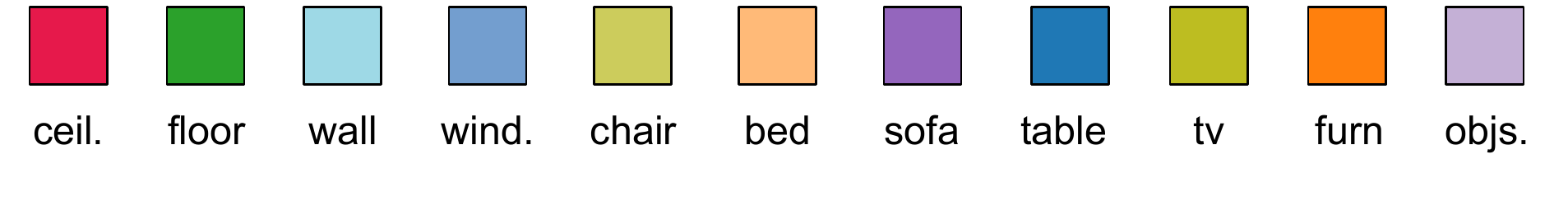}%    
  \caption{
	  \small{Qualitative results on NYUv2 Kinect. From left to right: Input RGB-D image, ground truth, result obtained by \cite{song2016ssc}, and result obtained by our approach. Overall, our completed semantic 3D scenes are less cluttered and show a higher voxel class accuracy compared to \cite{song2016ssc}.}
	  }
  \label{fig:PixelPredQualRes}
\end{figure*}

\section{Conclusions}
In this work, we have proposed a two stream approach for 3D semantic scene completion. In contrast to previous works, the proposed approach leverages depth and semantic information of the visible part of the scene for this task. In our experiments, we have shown that the proposed three-channel encoding for the semantic volume is not only memory efficient but it also results in higher accuracies compared to a single-channel encoding and is competitive to a memory expensive one-hot encoding. The proposed approach achieves state-of-the-art results for semantic scene completion on the NYUv2 dataset while also providing much faster inference times than approaches based of TSDF input features.

\section*{Acknowledgement}
A special thanks goes to Christian Grund for visualizing the qualitative results in Blender. The work has been funded by the Deutsche Forschungsgemeinschaft (DFG, German Research Foundation) – GA 1927/2-2 (FOR 1505 Mapping on Demand).
% \clearpage

% {\small
% \bibliographystyle{ieee}
% \bibliography{egbib}
% }

% New Bibstyle:
{\small
\bibliographystyle{ieee_fullname}
\bibliography{egpaper_for_review}
}

\end{document}